\documentclass{article}
\usepackage{float}

\usepackage{arxiv}

\usepackage[utf8]{inputenc} % allow utf-8 input
\usepackage[T1]{fontenc}    % use 8-bit T1 fonts
\usepackage{hyperref}       % hyperlinks
\usepackage{url}            % simple URL typesetting
\usepackage{booktabs}       % professional-quality tables
\usepackage{amsfonts}       % blackboard math symbols
\usepackage{nicefrac}       % compact symbols for 1/2, etc.
\usepackage{microtype}      % microtypography
\usepackage{lipsum}
\usepackage{graphicx}
\graphicspath{ {./images/} }
\usepackage{adjustbox}
\usepackage{enumerate}
\usepackage{multirow}
\usepackage{float}

\title{Local Herb Identification Using Transfer Learning: A CNN-Powered Mobile Application for Nepalese Flora}

\author{
 Prajwal Thapa, 
  \hspace{1mm} Mridul Sharma, 
 \hspace{1mm} Jinu Nyachhyon, 
 \hspace{1mm} Yagya Raj Pandeya\\
  \\
  }

\begin{document}
\maketitle
\begin{abstract}
Herb classification presents a critical challenge in botanical research, particularly in regions with rich biodiversity such as Nepal. This study introduces a novel deep learning approach for classifying 60 different herb species using Convolutional Neural Networks (CNNs) and transfer learning techniques. Using a manually curated dataset of 12,000 herb images, we developed a robust machine learning model that addresses existing limitations in herb recognition methodologies. Our research employed multiple model architectures, including DenseNet121, 50-layer Residual Network (ResNet50), 16-layer Visual Geometry Group Network (VGG16), InceptionV3, EfficientNetV2, and Vision Transformer (VIT), with DenseNet121 ultimately demonstrating superior performance.  Data augmentation and regularization techniques were applied to mitigate overfitting and enhance the generalizability of the model. This work advances herb classification techniques, preserving traditional botanical knowledge and promoting sustainable herb utilization.
\end{abstract}

% keywords can be removed

\section{Introduction}
Herbs have been used for centuries for their medicinal, culinary, and aromatic properties. They play a significant role in various cultures and traditional systems of medicine. Herbs are known for various properties like medicinal properties, natural remedies, culinary enhancement, nutritional values, aromatherapy, environmental benefits, and cultural and traditional significance. Herbs and related products are commonly used by patients who seek conventional health care. Some herbs have been extensively studied, but little is known about others \cite{winslow1998herbs}. Due to the increasing interest in herbal medicines, there is a need that arise to high standards of high-quality control of herbs. There is a huge gap in the recognition of the herbs that are all around us for the modern generation, who are not very familiar with the useful herbs. In this paper, we are going to introduce a deep learning-enabled mobile application that can run entirely on smartphones for the efficient and effective herb image recognition in resource-limited situations. The motivation for developing a mobile application for the classification of invasive herbs stems from the significant threat these plants pose to ecosystems and the need for efficient management strategies. By providing a user-friendly tool accessible on smartphones, the application empowers individuals to contribute to the identification and management of invasive species. The increasing accessibility of smartphones and the convenience they offer make a mobile application an ideal platform for real-time identification and reporting. Incorporating machine learning algorithms or image recognition techniques enables users to identify invasive herbs based on visual cues, allowing even those with limited botanical knowledge to participate actively. The goal is to engage a wide range of users, facilitate rapid identification, and support conservation efforts aimed at addressing the detrimental effects of invasive herbs on native plant species and biodiversity. The main goal of this mobile application is to help users learn about herbs effortlessly, without requiring prior knowledge, by leveraging machine learning. The steps involved while building this application are data preparation, training the collected data in various model architectures, mobile application framework selection and comparisons, and choosing the best-performing model so that we can integrate that particular model into our mobile application. Collecting the data on the invasive plants is quite challenging. We collected the data by crawling the internet and added some manually captured photos of the plant, which is further discussed in \nameref{data} sub-section. We trained the collected data in various available model architectures and selected the best-performing model. We made a simple but user-friendly user interface in the Flutter framework and integrated the best model for our use case.
The main challenges of the classification of invasive herb plants involve the task of distinguishing and categorizing different types of invasive herbaceous plants based on their characteristics and features. Invasive herbaceous plants are species that are non-native to a particular ecosystem and have a tendency to spread aggressively, potentially causing harm to the environment, economy, or human health.

The goal of invasive herb plant classification is to develop a model or system that can accurately identify and classify these invasive plant species. This typically involves using Machine Learning or Deep Learning algorithms to analyze and extract relevant features from images or other data sources associated with the plants. The current trend of identifying the herb plants and extracting their information may require expertise on the herbs and botanical knowledge. This may create a knowledge gap in the upcoming generation. Without knowing the importance and knowledge of the important herbs, it may lead to the extinction of the plants. In manual recognition, one popular method is herb fingerprint technology. It extracts the biological fingerprint of herbs by complicated chemical steps, including herb splitting, gas chromatograph-mass spectrometer-computer (GC-MS) analysis, and fingerprint generation, and then the fingerprint is compared to the template of each category based on their similarity. Due to its reliability and accuracy in herb recognition, it has been accepted as a standard pipeline worldwide. Another solution for manual recognition is the experience-based method, where professionals use their long-term experiences by looking, smelling, tasting, touching, and hearing. However, manual recognition has its limitations. Firstly, it cannot work without professionals with rich knowledge and experience, or chemical materials and devices that may not be available for resource-limited settings. Secondly, it will cost too many resources and time to recognize thousands of herbs manually \cite{sun2022deep}. Recently, a systematic literature review by Wäldchen et al. has surveyed computer vision-based plant species identification techniques, illustrating the growing shift from manual to automated plant recognition in scientific research~\cite{wäldchen2018plant}.

\begin{figure}[htbp]
  \centering
  \includegraphics[width=\columnwidth]{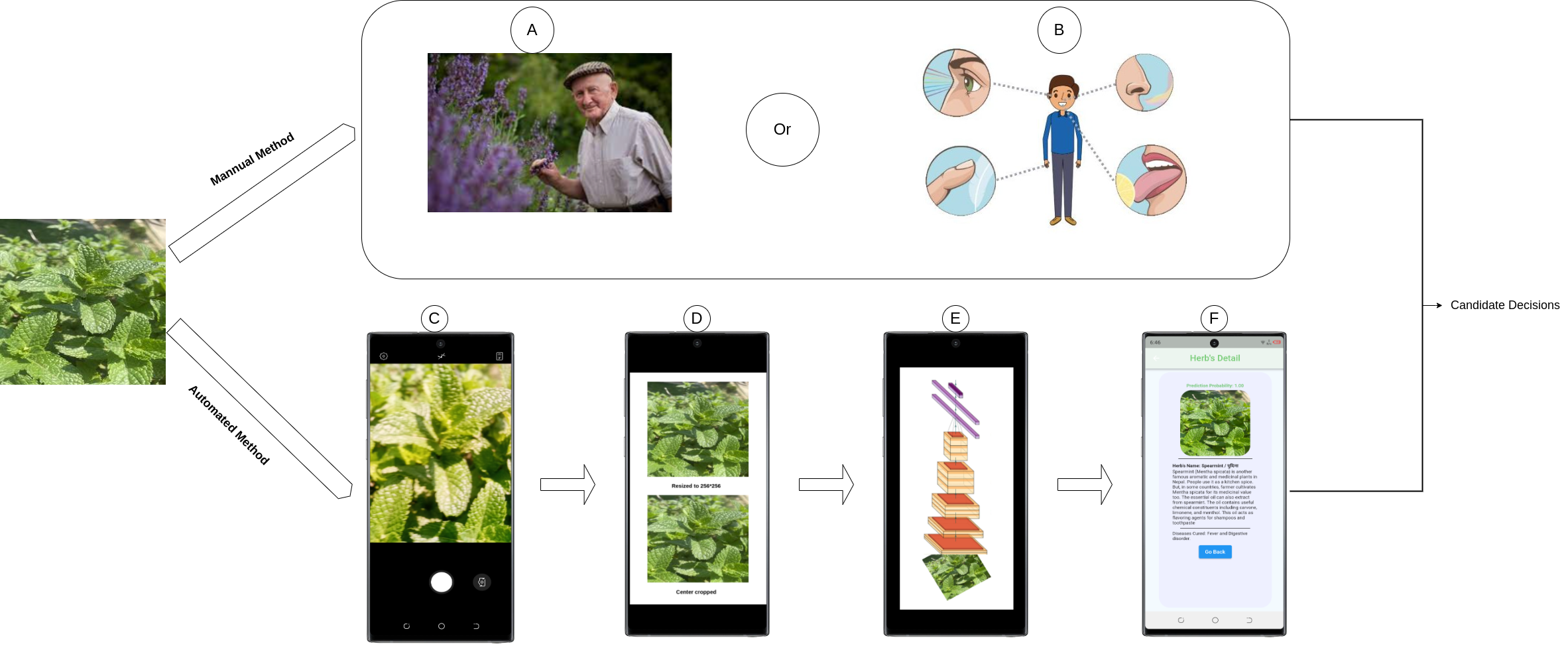}
  \caption{Manual and automatic herb recognition}
  \label{mannual}
\end{figure}

Traditional way of recognition of herbs\ref{mannual}(A) includes an experience\ref{mannual}(B) over the herbs like an expert knowledge or by using some knowledge of the herbs and using our sense organs like tasting, smell, sight, and touch \ref{mannual}(C).

The gap we are going to fill here is that  Users will be able to effortlessly capture images of herbs using their smartphones, and the app will instantly provide reliable results. Seamless integration with other tools, such as gardening or health applications, will enhance its versatility, catering to a wide range of user needs. Moreover, the application will act as an extensive educational platform, offering valuable insights into each identified herb's uses, medicinal properties, and optimal growing conditions. With less knowledge, just operating the application as shown in \ref{mannual}(F) will provide the user with the information about the herbs in front of them.

In the automatic herb recognition procedure, researchers first utilize herb images and process them into their low-level features for recognition. Chen and colleagues\cite{chen2020image} developed a color matching template that incorporates two observation surfaces of herb pieces. They demonstrated that this template exhibits some resistance to interference in rotation, shape, and color. On the other hand, Liu et al.\cite{liu2013image} employed gray-scale images derived from color images to extract texture features. They utilized the gray-level co-occurrence matrix for recognition purposes. To differentiate herbs with similar appearances, Ming et al.\cite{ming2019rapid} combined Raman spectroscopy with the Support Vector Machine. However, these studies primarily focused on a limited number of herb categories. This limitation arose from the fact that low-level properties were unable to handle the substantial variations present in many categories. All of this work has been developed in foreign ecosystems and fails to address the specific challenges of the Nepalese context, particularly the absence of local herbs found in Nepal. Many research investigations have relied on the utilization of Deep Neural Networks (DNNs), which are coupled with robust and costly computational hardware to facilitate the computational requirements. This combination is frequently employed within medical facilities and laboratories, with the intention of achieving a high level of system automation and ensuring system stability. Most of the work done in the herbs and plant detection methods has been done in the harvested datasets and are most useful in the lab or for chemical procedures; classification done in the wild herb plants is rare. Another main goal of this application is to address the lack of recognized work on the classification of invasive herbs, particularly within Nepal. As we know, Nepal is very rich in biodiversity. Herbs have played a significant role in Nepal's culture, tradition, and everyday life for centuries. The country's diverse geography, ranging from the plains of the Terai to the towering Himalayas, fosters a rich variety of flora, including numerous medicinal and aromatic herbs. Recent advances in Deep Learning have enabled robust mobile applications for plant disease detection and diagnosis, as shown by Ferentinos~\cite{ferentinos2018deep}. This supports the growing trend of employing Deep Neural Networks (DNN) for efficient in-field plant identification.

\section{Materials and Methods}
\subsection{Data}
\label{data}
In this study, we have manually labeled datasets of approximately 12000 images. Further data collection and preprocessing are discussed below:
\subsubsection{Data Collection}
The data we gathered is mainly from three sources. We manually captured the available herb plants in our locality, collected the available herbs datasets from Kathmandu University AI Lab archives, and collected the herbs available in Nepal by crawling the internet. The purpose of image processing when dealing with herb images is to serve two primary objectives. Firstly, it involves preparing the image for computation using a DNN. The image size can vary significantly due to variations in hardware and resolution. However, the DNN requires a fixed input image size, such as 256x256 pixels, throughout our evaluation. Hence, it becomes necessary to process the image to match the target size, a rule that applies to both the training and testing phases. Secondly, image processing serves as an effective method for data augmentation, enhancing the robustness of recognition. Recent surveys show image data augmentation to be critical in achieving good generalization from limited real-world datasets~\cite{shorten2019survey}. In our study, we trained and evaluated our method using the dataset we created, consisting of 60 herb categories and 12000 herb images. Some classes with their sample images are discussed in Table \ref{collection}. Training the DNN solely on this dataset without any data augmentation would easily lead to overfitting, compromising the generality of herb image recognition. It is important to note that this approach is mainly applied during the training phase, where techniques like augmentation are used, while the processing during the testing phase remains relatively straightforward.

\begin{table}[ht]
  \centering
  \begin{adjustbox}{width=1\textwidth}
  \begin{tabular}{cccc}
    \hline
    \textbf{Class/label} & \textbf{Train set} & \textbf{Validation}  & \textbf{Test set}\\
    \hline
    Psidium guajava & \includegraphics[width=2cm]{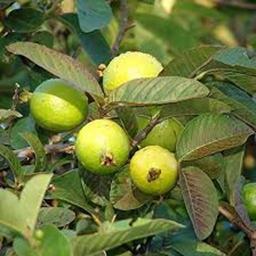} \includegraphics[width=2cm]{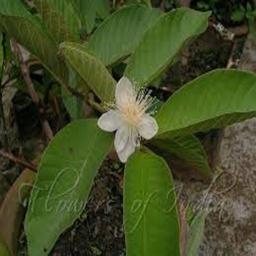}  & \includegraphics[width=2cm]{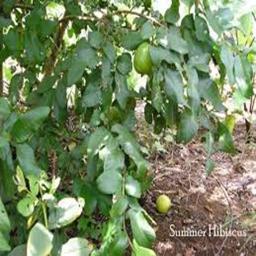} \includegraphics[width=2cm]{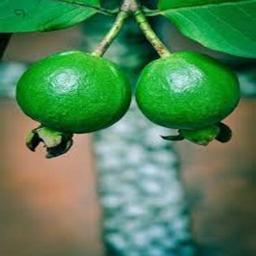}  & \includegraphics[width=2cm]{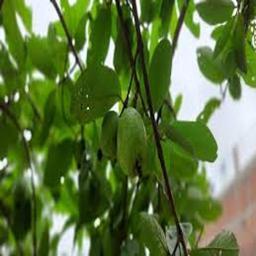} \includegraphics[width=2cm]{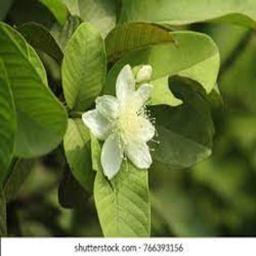}\\
    \hline
    Rauwolfia serpentina & \includegraphics[width=2cm]{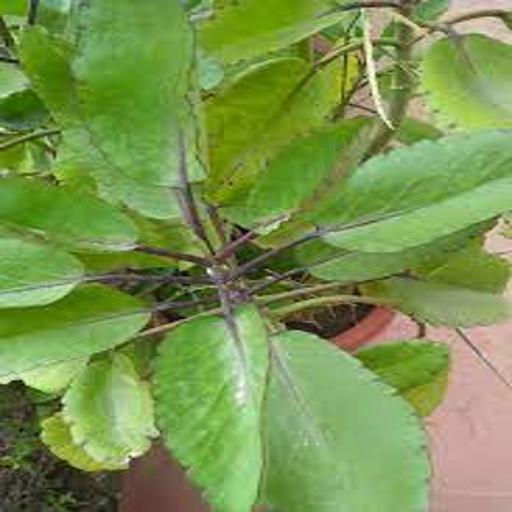} \includegraphics[width=2cm]{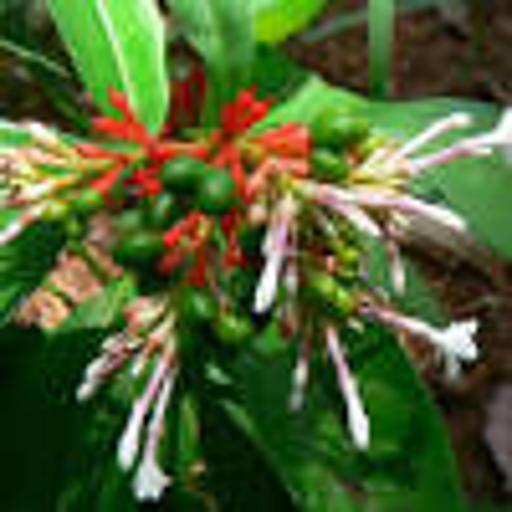}  & \includegraphics[width=2cm]{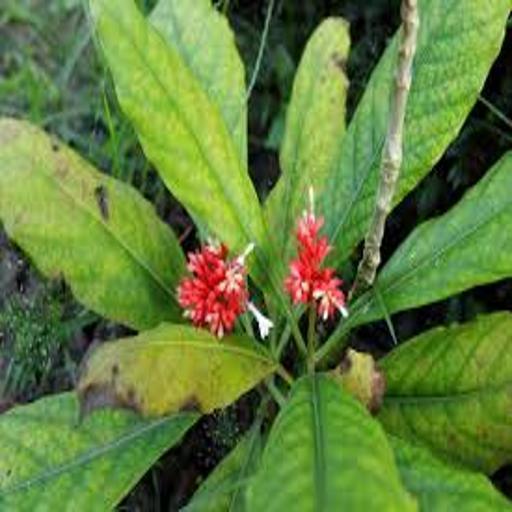} \includegraphics[width=2cm]{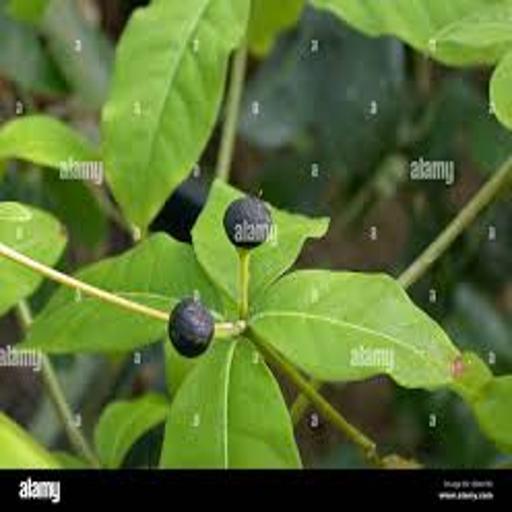}  & \includegraphics[width=2cm]{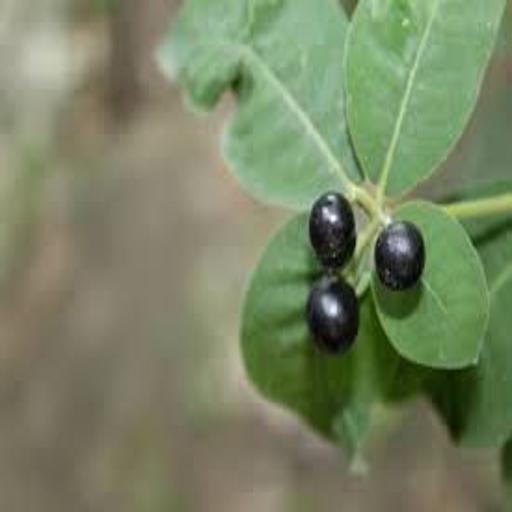} \includegraphics[width=2cm]{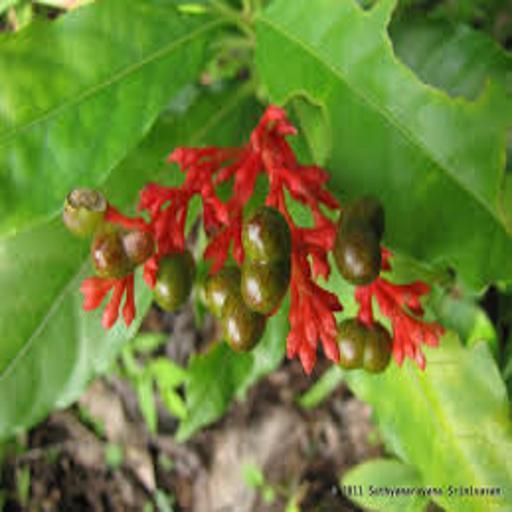}\\
    \hline
    Mentha spicata & \includegraphics[width=2cm]{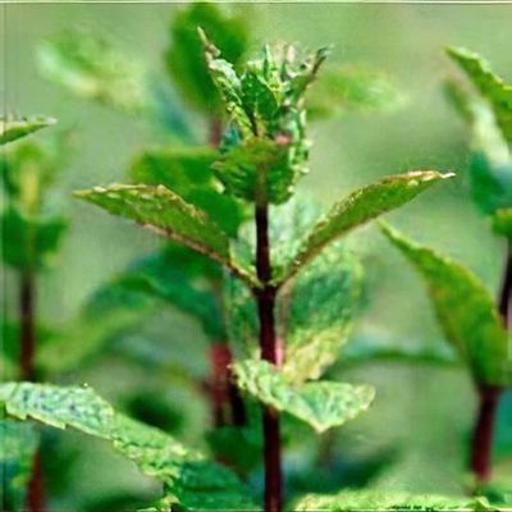} \includegraphics[width=2cm]{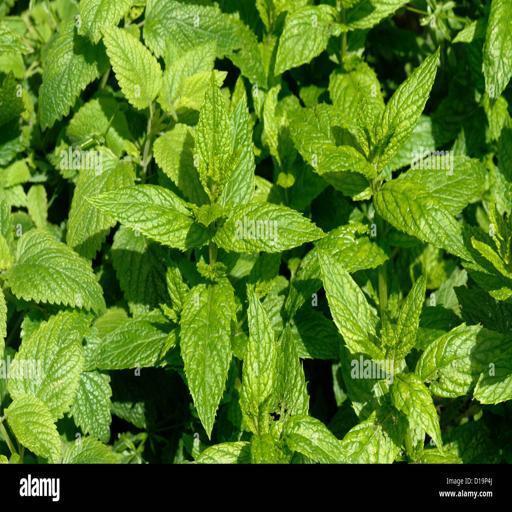}  & \includegraphics[width=2cm]{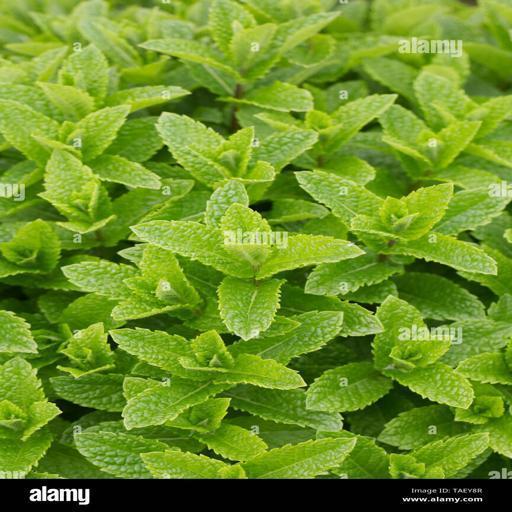} \includegraphics[width=2cm]{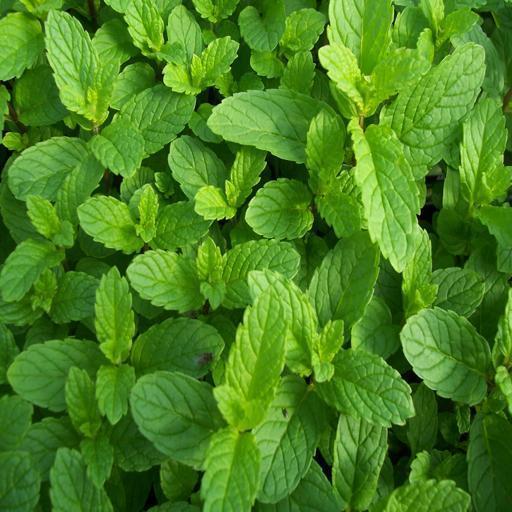}  & \includegraphics[width=2cm]{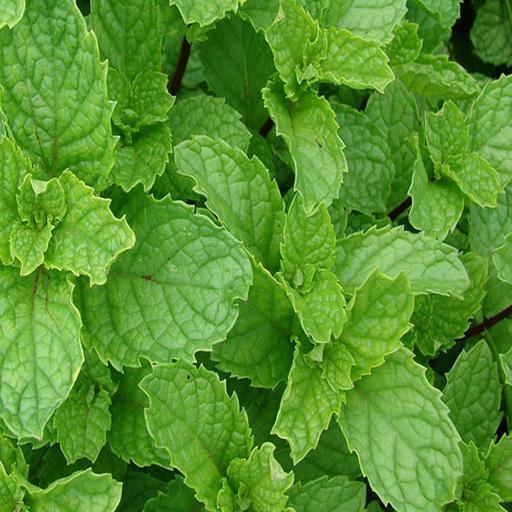} \includegraphics[width=2cm]{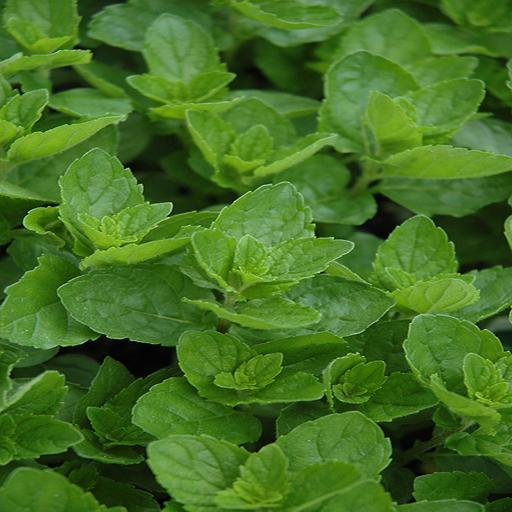}\\
    \hline \\
  \end{tabular}
  \end{adjustbox}
  \caption{Few sample images related to some classes on various datasets}
  \label{collection}
\end{table}

\subsubsection{Data Cleaning and Preprocessing}
To get the sound data and a dataset of our interest, we went through various data cleaning and preprocessing methods:
\begin{enumerate}[1.]
  \item Cleaning of low-quality images to get equivalent aspect ratio, filtering the images with our interest points
  \item Resized the images to a size of 256*256 to get sound resolutions
  \item Normalized the image's pixel values within the range [0,1]
\end{enumerate}

\subsubsection{Data Labeling}
Data labeling is a crucial step in preparing a dataset for herb classification. It involves assigning appropriate labels or categories to each data point, specifically the 60 different herbs in this case. Accurate and consistent labeling is essential to ensure that the machine learning model can learn the correct patterns and accurately classify the herbs based on their features. Our task involved multi-class classification with 60 distinct classes; for simplicity and consistency, we labeled our data using the scientific names of each herb as their respective class labels. We collected the scientific names of each herb and annotated them to their image directories. Furthermore, we gathered information about each herb, such as its characteristics and medicinal uses, from online sources. This was essential because, after identifying the herb during data labeling, we needed to provide relevant details alongside its name. We mapped this information to each corresponding label to ensure a complete and informative output.

\subsubsection{Datasets Creation}
After cleaning and preprocessing the data, we label them. Then, we created three mutually exclusive datasets: train, validation, and test datasets as shown in Table \ref{distribution}. For the fair division of the datasets, we have randomly shuffled the images before creating the mentioned datasets. The number and ratio among the datasets are further discussed in the table. 

\begin{table}[ht]
  \centering
  \label{distribution}
  \begin{tabular}{ccccc}
    \hline
    \textbf{Classes} & \textbf{total images} & \textbf{train data} & \textbf{validation data}& \textbf{test data}\\
    \hline
    60 classes & 12000 & 9000 & 1500 & 1500 \\
   
    \hline \\
  \end{tabular}
  \caption{Distribution of the data among the datasets}
\end{table}

\subsection{Transfer Learning}
Due to the limitation on the data, as we do not have sufficient herb images, to obtain the optimal results in our application, we used a transfer learning mechanism. Transfer learning aims at improving the performance of target learners on target domains by transferring the knowledge contained in different but related source domains. In this way, dependence on large amounts of target domain data can be reduced when constructing target learners. Due to the wide application prospects, transfer learning has become a popular and promising area in machine learning \cite{zhuang2019comprehensive}. Here we used two approaches of transfer learning, first we used the ImageNet ~\cite{deng2009imagenet} weights in our models, secondly we trained a huge plant dataset to extract the plant features of about 80-90 thousands of plant species image data. Later, those trained data's weights are transferred to our 60-herb classification model to obtain better performance.

\subsection{Regularization}
Regularization in the context of Convolutional Neural Networks (CNNs) refers to a set of techniques that aim to prevent overfitting and improve the generalization ability of the model. Overfitting occurs when a CNN performs extremely well on the training data but fails to generalize well to unseen data. Regularization techniques introduce additional constraints or penalties to the CNN's loss function during training, discouraging complex or overly complex models that may be more prone to overfitting. These techniques help to control the model's capacity and prevent it from memorizing the training data without capturing the underlying patterns. The problems of model overfitting and instabilities are all due to the lack of enough data. Obtaining more data is not always possible, so to increase the number of labeled data points, we utilized the data augmentation method. To mitigate the common issue of overfitting in deep learning models, a combination of four well-known techniques was employed. These techniques include L2 regularization, dropout 
 \cite{Srivastava2014DropoutAS}, batch normalization layer \cite{Ioffe2015BatchNA}, and data augmentation \cite{Perez2017TheEO}. These methods collectively aim to reduce overfitting and improve the generalization ability of the models. As shown in Table \ref{data_augmentation}, we use various augmentation methods to increase the number of data points.
 \begin{table}[ht]
  \centering
  
  \label{data_augmentation}
  \begin{tabular}{cc|cc}
    \hline
    \textbf{Method} & \textbf{Value} & \textbf{Method} & \textbf{Value}\\
    \hline
    Flip upside down & 0.5 probability & Flip horizontal & 0.5 probability \\
    Rotate image & -45 and 45 degree & Gaussian noise & scale(0,1 * 255) \\
    Multiply the pixel value & by random between (0.8,1.2) & Crop & random centre \\
    Elastic transformation & between 0.5 to 3.5\\
    \hline \\
  \end{tabular}
  \caption{Data augmentation methods and their corresponding values}
\end{table}
\vspace{-3em}

\section{Experimental Setup and Evaluation}
\subsection{Network Architecture}
Convolutional Neural Networks (CNNs) are a class of DNN designed specifically to process and analyze visual data, particularly images. CNNs have revolutionized the field of computer vision by achieving state-of-the-art performance on a wide range of tasks, including image classification, object detection, and image segmentation. At the core of CNNs lies the concept of exploiting the inherent spatial structure present in images. Unlike traditional neural networks, which treat input data as a flat vector, CNNs preserve the spatial relationships between pixels by leveraging convolutional layers. These layers employ learnable filters, known as kernels or feature detectors, to perform localized filtering operations on the input. The architecture of a typical CNN comprises multiple types of layers, each serving a distinct purpose. Convolutional layers, the fundamental building blocks, consist of a set of learnable filters. These filters convolve with the input image, producing a set of feature maps. By sliding these filters across the spatial dimensions of the input, convolutional layers capture local patterns and features, facilitating the learning of hierarchical representations.

\begin{figure}[htbp]
  \centering
  \includegraphics[width=\columnwidth]{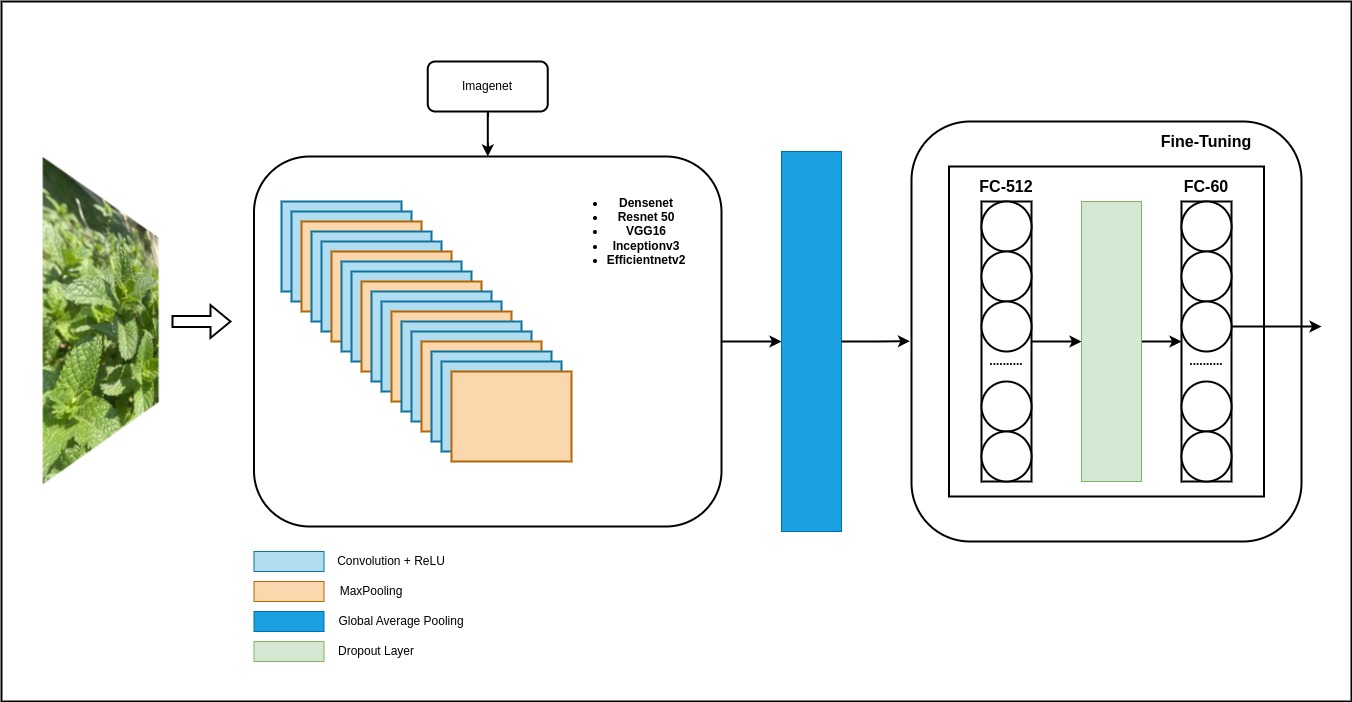}
  \caption{Overall architecture of our model used}
  \label{CNN}
\end{figure}

For our problem statement, not to be focused on a particular architecture, we experimented with our datasets through various model architectures like DenseNet121, ResNet50, VGG16, InceptionV3, EfficientNetV2, and VIT. We evaluate the performance of each model architecture to pick the best-performing model for our data. We use these as a base model pretrained with the ImageNet weights. Then the work of the pooling layer is to flatten the output of the base model and the activation fully connected layer after the pooling layer is replaced by 60 nodes, as we have 60 various classes of herbs. With the softmax activation, we finally predict the highest probability of the herb in the final layer as shown in Figure \ref{CNN}. The DenseNet architecture, which emphasizes dense connections between layers, was presented by Huang et al. and has shown great effectiveness in a variety of computer vision tasks~\cite{huang2017densely}.

\subsection{Evaluation}
\vspace{-1em}
The content discusses the importance of evaluation in machine learning model performance assessment for herb classification. It highlights the use of evaluation metrics to measure the accuracy of the model in predicting class labels for unseen herb images in a 60-class classification task. The cost function, or loss function, is described as a crucial element in training the model by measuring the disparity between predicted outputs and ground truth labels. The categorical cross-entropy loss is mentioned as a common choice for the cost function in the context of 60-class herb classification. Various performance measures and metrics are emphasized, including accuracy, precision, recall, F1-score, and the confusion matrix, which aid in evaluating the effectiveness and efficiency of the machine learning models. Additionally, the AUC score is mentioned as another measure used in the experiment.
\vspace{-2em}

\section{Results}
\vspace{-2em}
After training our datasets using different model architectures and employing transfer learning and fine-tuning techniques, we assessed the performance of each model. As expected, the accuracy improved significantly after applying transfer learning. Our datasets exhibited commendable performance across all models, with DenseNet121 and InceptionV3 showing particularly promising results. The comprehensive evaluation metrics, including AUC score and F1 score, were computed exclusively on the test data and are presented in the Table \ref{results}. The AUC score measures the model's ability to distinguish between classes, while the F1 score balances precision and recall. After careful analysis, we found that the DenseNet121 model outperformed others, displaying exceptional performance in the classification of the 60 herb classes. Thus, we decided to select DenseNet121 for further use cases and evaluations. For a more detailed assessment of the selected model, please refer to \ref{subsec:densenet_evaluation}.

\begin{table}[ht]
  \centering
  \label{results}
  \begin{tabular}{c|c|cccc}
  \hline
  \multicolumn{1}{c|}{\textbf{}} &\textbf{} & \multicolumn{4}{c}{\textbf{Metrics}} \\

  \textbf{Models} & \textbf{Dataset} & \textbf{Accuracy} & \textbf{Loss} & \textbf{Auc Score} & \textbf{F1 Score} \\
  \hline
  \multicolumn{1}{c|}{\textbf{DenseNet121 ~\cite{huang2017densely}}} &\textbf{Train} & 95.90 & 0.14 &\multicolumn{1}{c}{\multirow{2}{*}{0.88}}  &\multicolumn{1}{c}{\multirow{2}{*}{0.78}} \\
  \multicolumn{1}{c|}{\textbf{}} &\textbf{Validation} & 82.64 & 0.78 &  &  \\
  \hline
  \multicolumn{1}{c|}{\textbf{ResNet50 ~\cite{he2016deep}}} &\textbf{Train}& 98.02 & 0.06 &\multicolumn{1}{c}{\multirow{2}{*}{0.83}}  &\multicolumn{1}{c}{\multirow{2}{*}{0.68}} \\
  \multicolumn{1}{c|}{\textbf{}} &\textbf{Validation} & 69.22 & 0.98 &  &  \\
  \hline
  \multicolumn{1}{c|}{\textbf{VGG16 ~\cite{simonyan2015very}}} &\textbf{Train} & 71.90 & 0.93 &\multicolumn{1}{c}{\multirow{2}{*}{0.83}}  &\multicolumn{1}{c}{\multirow{2}{*}{0.68}} \\
  \multicolumn{1}{c|}{\textbf{}}&\textbf{Validation} & 69.40 & 1.15 & & \\
  \hline
  \multicolumn{1}{c|}{\textbf{Inceptionv3 ~\cite{szegedy2016rethinking}}} &\textbf{Train}& 85.55 & 0.44 &\multicolumn{1}{c}{\multirow{2}{*}{0.86}}  &\multicolumn{1}{c}{\multirow{2}{*}{0.74}} \\
  \multicolumn{1}{c|}{\textbf{}} &\textbf{Validation}& 74.95 & 1.10 & & \\
  \hline
  \multicolumn{1}{c|}{\textbf{EfficientNetv2 ~\cite{tan2019efficientnet}}} &\textbf{Train}& 97.43 & 0.08 &\multicolumn{1}{c}{\multirow{2}{*}{0.67}}  &\multicolumn{1}{c}{\multirow{2}{*}{0.53}} \\
  \multicolumn{1}{c|}{\textbf{}} &\textbf{Validation}& 62.80 & 2.02 & & \\
  \hline
  \multicolumn{1}{c|}{\textbf{VIT ~\cite{dosovitskiy2021image}}} &\textbf{Train}& 97.99 & 0.06 &\multicolumn{1}{c}{\multirow{2}{*}{0.57}}  &\multicolumn{1}{c}{\multirow{2}{*}{0.52}} \\
  \multicolumn{1}{c|}{\textbf{}} &\textbf{Validation}& 46.24 & 4.06 & & \\
  \hline 
  \end{tabular} 
  \vspace{1em}
\caption{Results in various models along with their metrics}
\end{table}
\vspace{-4em}
%****************************************************************
%\section{Conclusions}

\subsection{Evaluation on DenseNet121} \label{subsec:densenet_evaluation}
\vspace{-1em}
According to the results presented in the Table \ref{results}, the DenseNet121 model exhibits exceptional performance on our dataset. Among the various model architectures evaluated, DenseNet121 consistently achieves outstanding performance across multiple performance metrics. Notably, it demonstrates remarkably high accuracy, AUC score, and F1 score, surpassing the performance of all other model architectures. The confusion matrix is the Table \ref{dense_confusion} used in herb classification to evaluate the performance of a DenseNet121 model. It presents a comprehensive view of the model's predictions by counting the true positives, true negatives, false positives, and false negatives. By organizing the information into rows and columns, the confusion matrix helps in analyzing the model's strengths and weaknesses. The matrix's diagonal elements represent correctly classified instances, while the off-diagonal elements indicate misclassifications, providing valuable insights into the model's accuracy and error patterns. Given its consistently strong performance, DenseNet121 is an ideal choice for further implementation and utilization in our project. 

\begin{table}[H]
  \centering
  \includegraphics[height=10cm]{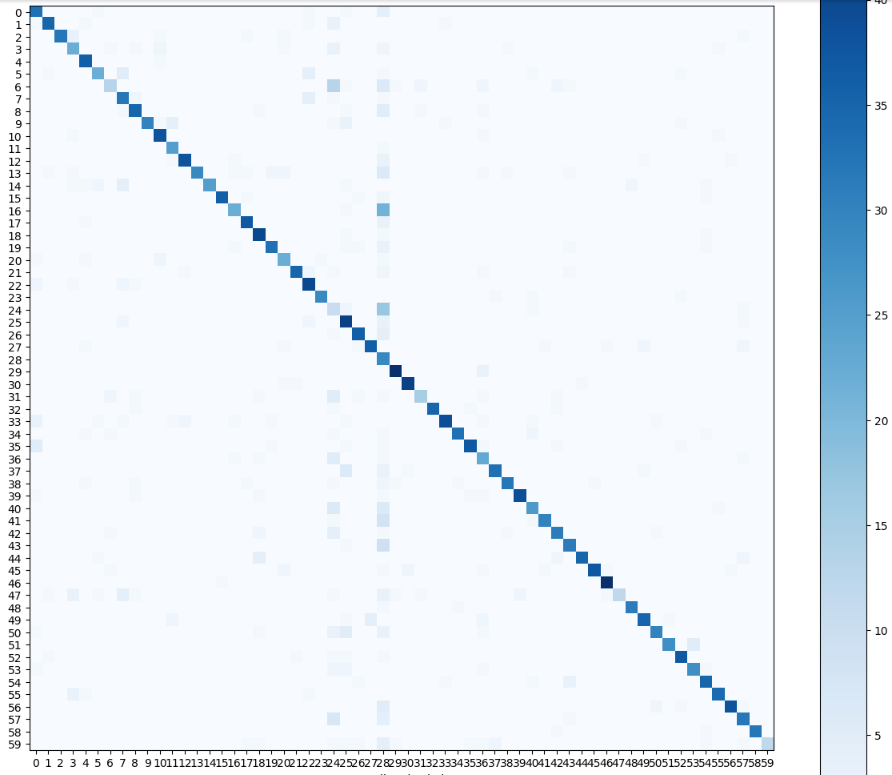}
  \caption{Confusion matrix evaluated on the validation data on 1500 images}
  \label{dense_confusion}
\end{table}

\begin{figure}[H]
  \centering
  \includegraphics[width=\columnwidth]{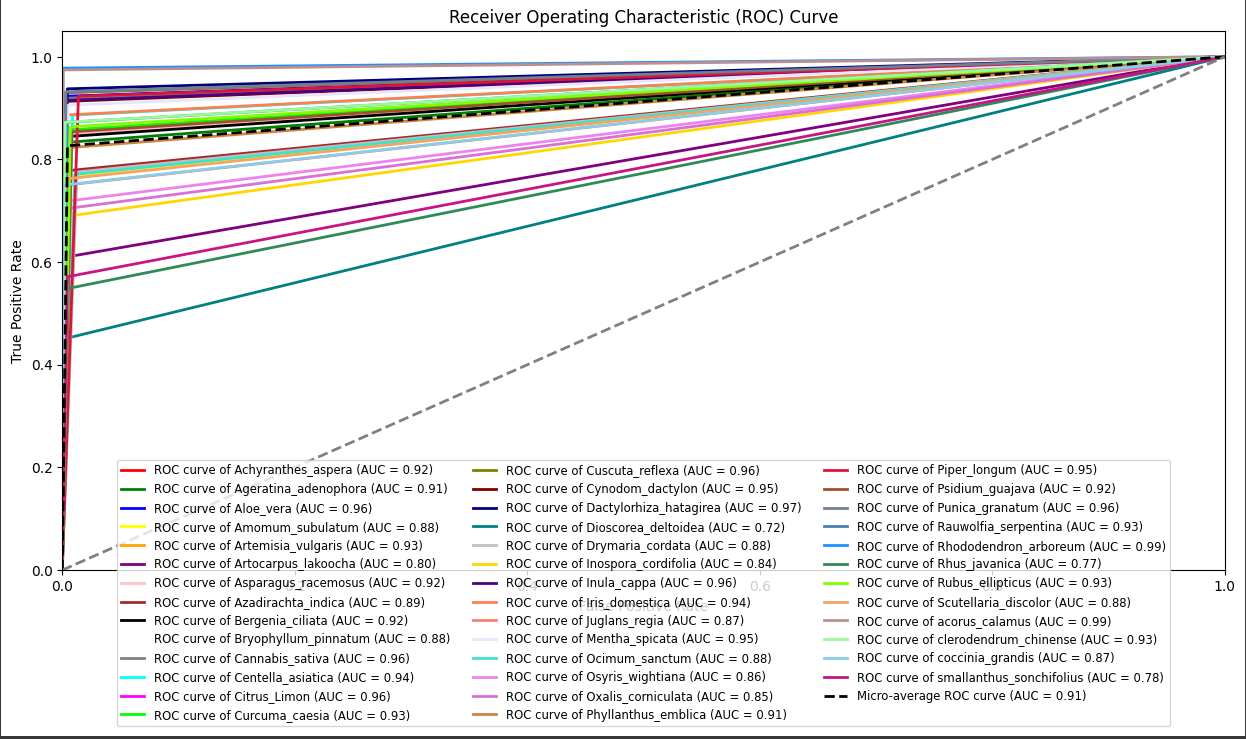}
  \caption{AUC score of the model with each AUC score of 60 different herb species}
  \label{dense_auc}
\end{figure}

The ROC curve [\ref{dense_auc}] is a graphical representation of the performance of a binary classification model. It illustrates the trade-off between the true positive rate (sensitivity) and the false positive rate (1 - specificity) for various classification thresholds. However, in the case of multi-class classification like 60 herbs classification, the ROC curve is typically computed by considering each class against the rest (one-vs-all approach) or by using techniques like micro-averaging or macro-averaging. The area under the ROC curve (AUC) is commonly used as a metric to quantify the overall performance of the model. A higher AUC indicates better discrimination ability, with a perfect classifier having an AUC of 1. 

\begin{figure}[ht]
  \centering
  \includegraphics[width=\columnwidth]{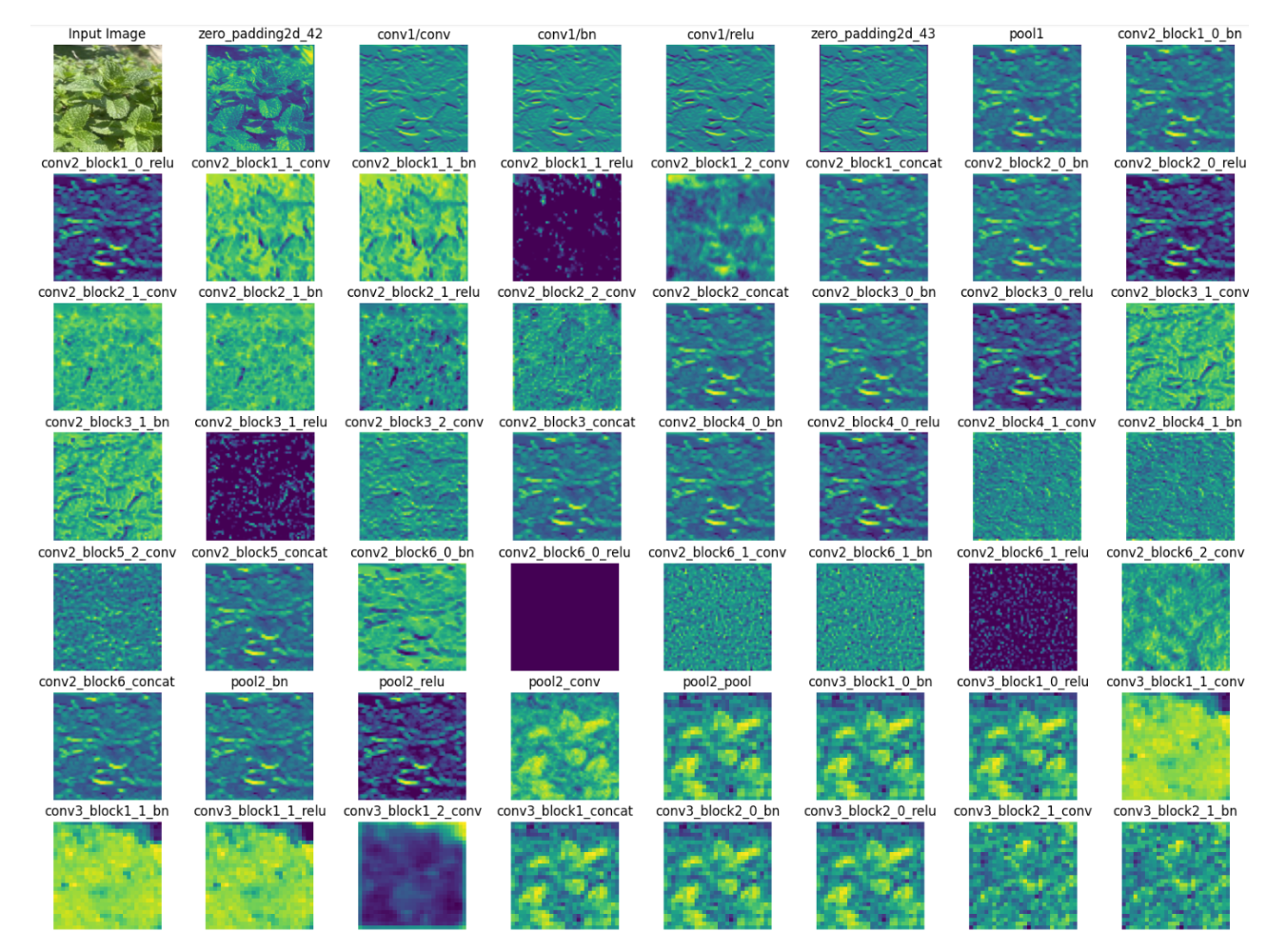}
  \caption{Features learned by our model in each layer}
  \label{features_learned}
\end{figure}

Feature extraction techniques are applied to extract relevant information or characteristics from herb images. These techniques aim to capture distinctive patterns or features that can discriminate between different herb classes. While we pass our herb data through our model, our model learns some features through the filters. To understand how features are learned from the image data, we visualized the outputs of each layer in DenseNet121, as shown in Figure~\ref{features_learned}. Additionally, we used visualization techniques such as t-SNE and UMAP, following the approach of Draganov et al.~\cite{Draganov2023ActUpAA}, to further interpret and consolidate the high-dimensional features learned by the network.

\section{Network Implementation}
For the mobile implementation of the network, we chose the Flutter framework due to its versatility and robust cross-platform development capabilities. Flutter allows us to develop a single codebase that runs seamlessly on both iOS and Android platforms, dramatically reducing development time, effort, and maintenance costs compared to building native apps separately. This cross-platform efficiency is especially valuable for a herb classification application, where ensuring accessibility and uniform functionality across devices is essential. Its rich widget library and extensive customization options enable the creation of a visually engaging, intuitive, and responsive user interface. This is crucial for the herb classification app, where users benefit from a simple workflow, from capturing or selecting an image to viewing clear, actionable results. Moreover, Flutter's "hot reload" feature significantly accelerates the development cycle by allowing developers to see real-time changes, which is particularly beneficial when fine-tuning both the user interface and integrating machine learning components.

To power the herb classification functionality, we integrated TensorFlow Lite (TFLite), which allows us to deploy deep neural network (DNN) models efficiently on mobile devices. TensorFlow Lite is optimized for low-latency inference and minimal resource usage, making it an ideal choice for on-device inference, even on mid-range smartphones. After training the deep learning model on a high-performance computing environment, the model is converted into the TensorFlow Lite format, which reduces model size and optimizes performance without significant loss of accuracy. 

\begin{figure}[htbp]
  \centering
  \includegraphics[width=\columnwidth]{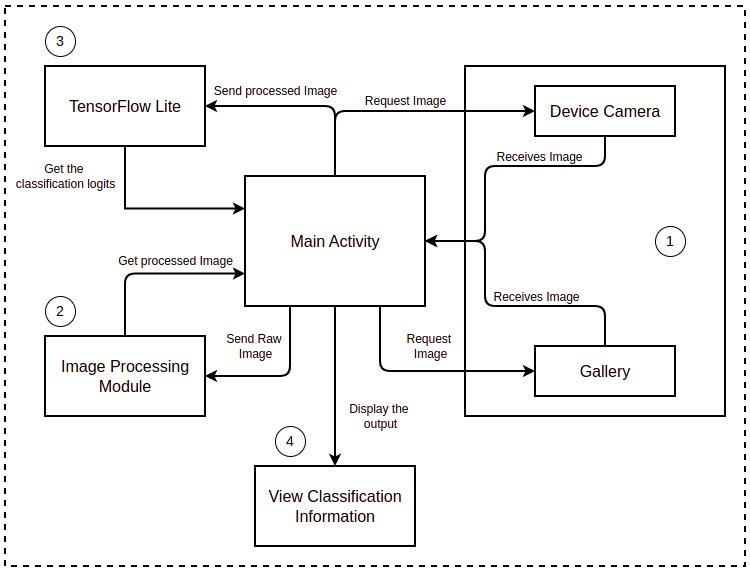}
  \caption{Activity diagram of the mobile application}
  \label{mobile_activity}
\end{figure}

In the Flutter application, the TFLite model is integrated using native platform channels or plugins, facilitating seamless communication between Dart code and the underlying machine learning engine. The app offers users two main input options: capturing a new herb image using the device camera or selecting an existing image from the gallery, as shown in Figure \ref{mobile_activity}. When the user chooses an image for classification, the Flutter app processes it locally using the pre-trained machine learning model. The model processes the input locally and returns a prediction that identifies the herb class. The app then displays not only the herb’s name but also supplemental information such as its medicinal uses, geographic distribution, and potential health benefits. This offline classification capability ensures that users can rely on the app even in remote areas with limited or no internet connectivity, a significant advantage in regions like Nepal, where rural access to technology and internet infrastructure can be inconsistent.

\section{Discussion and Analysis}
Classifying 60 herbs presents a myriad of challenges that require careful consideration and in-depth analysis. One of the primary hurdles lies in the inherent variability of herb characteristics. Each herb possesses unique features, including appearance, aroma, taste, chemical composition, and medicinal properties. Capturing and documenting this diversity in a comprehensive classification system is a complex task that demands not only technical precision but also a deep understanding of the intricacies of each herb’s biology and use.

Another significant challenge arises from the overlapping properties exhibited by certain herbs. For instance, distinguishing between herbs like thyme and oregano based solely on their appearance or flavor can be quite difficult, as they share many similar qualities. To ensure accurate classification, it becomes essential to examine a broad range of characteristics such as combining visual, chemical, and even genetic data. This multi-faceted approach ensures that subtle yet crucial differences are identified and appropriately recorded. Compounding this, the availability and reliability of information on all 60 herbs pose a major obstacle. Herb classification depends heavily on access to comprehensive data, including botanical descriptions, chemical compositions, traditional uses, and scientific research. Yet, some herbs may have limited research or insufficient documented data, impeding the classification process. Overcoming this challenge requires extensive research efforts and collaboration with botanists, chemists, herbalists, and other experts to bridge data gaps.

Cultural and regional variations further complicate herb classification. Different cultures and regions often have unique ways of categorizing and utilizing herbs, sometimes assigning different names or uses to the same species. What may be considered a single herb in one culture might be classified as multiple distinct herbs in another. Incorporating these cultural and regional perspectives into a cohesive and unified system requires a deep understanding of diverse practices and traditions. Moreover, the evolving nature of scientific knowledge presents an ongoing challenge in maintaining an up-to-date and accurate classification system. Research on herbs is continually progressing, with new discoveries, improved analytical techniques, and updated botanical information emerging regularly. Keeping pace with these advancements and integrating new findings demands regular updates and revisions, ensuring the system remains scientifically valid and relevant over time.

Despite these challenges, classifying 60 herbs offers numerous benefits and opportunities. One key advantage is the enhancement of medicinal and therapeutic applications. A well-structured classification system enables medical practitioners, herbalists, and researchers to understand the specific benefits, indications, contraindications, and potential side effects of different herbs. This facilitates the development of targeted treatments and therapies, ultimately improving patient health outcomes. Additionally, such a system holds significant culinary value. Chefs, culinary experts, and home cooks can use it to differentiate herbs based on their unique flavors, aromas, and culinary properties, allowing them to select and incorporate the most appropriate herbs into their recipes and enhance taste, presentation, and the overall dining experience.

Furthermore, accurate herb classification benefits agricultural practices by informing farmers and horticulturists of each herb’s specific needs, such as optimal soil types, climate preferences, and growth patterns. This supports optimized cultivation methods, resulting in higher yields and more efficient resource use. The classification of herbs also plays a crucial role in standardization and quality control within the herbal industry. A well-defined and universally recognized classification system allows regulatory bodies and manufacturers to implement rigorous quality control measures, helping authenticate herb identity, detect adulteration or substitution, and ensure consistency in product formulations. Ultimately, this safeguards consumer health and enhances trust in herbal products. Finally, classifying herbs contributes to the preservation of traditional knowledge. Documenting their properties, uses, and cultural significance ensures valuable information is passed down to future generations—preserving cultural heritage while promoting the sustainable use of natural resources and fostering a harmonious relationship between humans and the environment.

\section{Application Interface}

The mobile application developed based on the herb classification system enables users to easily identify herbs by capturing or uploading photos. It provides detailed information on each herb’s medicinal properties, culinary uses, and chemical composition through a clear and intuitive interface. The application features simple navigation with prominent buttons for taking photos or selecting images from the gallery, and displays classification results with high accuracy, including herb names, descriptions, and related health benefits. Designed with user accessibility in mind, the app ensures a smooth experience through organized layouts, readable text, and responsive interactions. Screenshots of the application are presented below:

\begin{figure}[H]
    \centering
    \begin{minipage}[b]{0.45\textwidth}
        \centering
        \includegraphics[width=\textwidth]{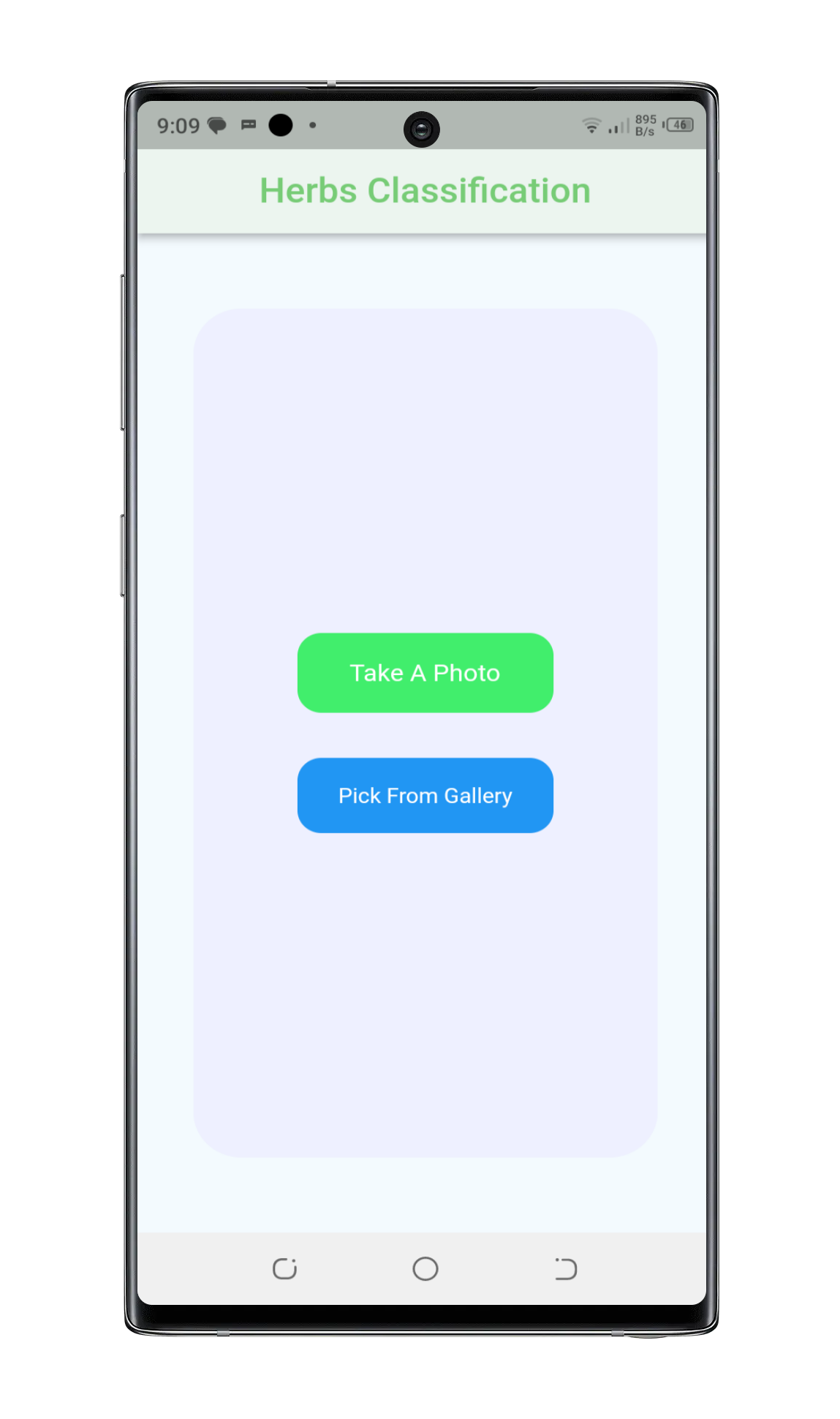}

    \end{minipage}
    \hfill
    \begin{minipage}[b]{0.45\textwidth}
        \centering
        \includegraphics[width=\textwidth]{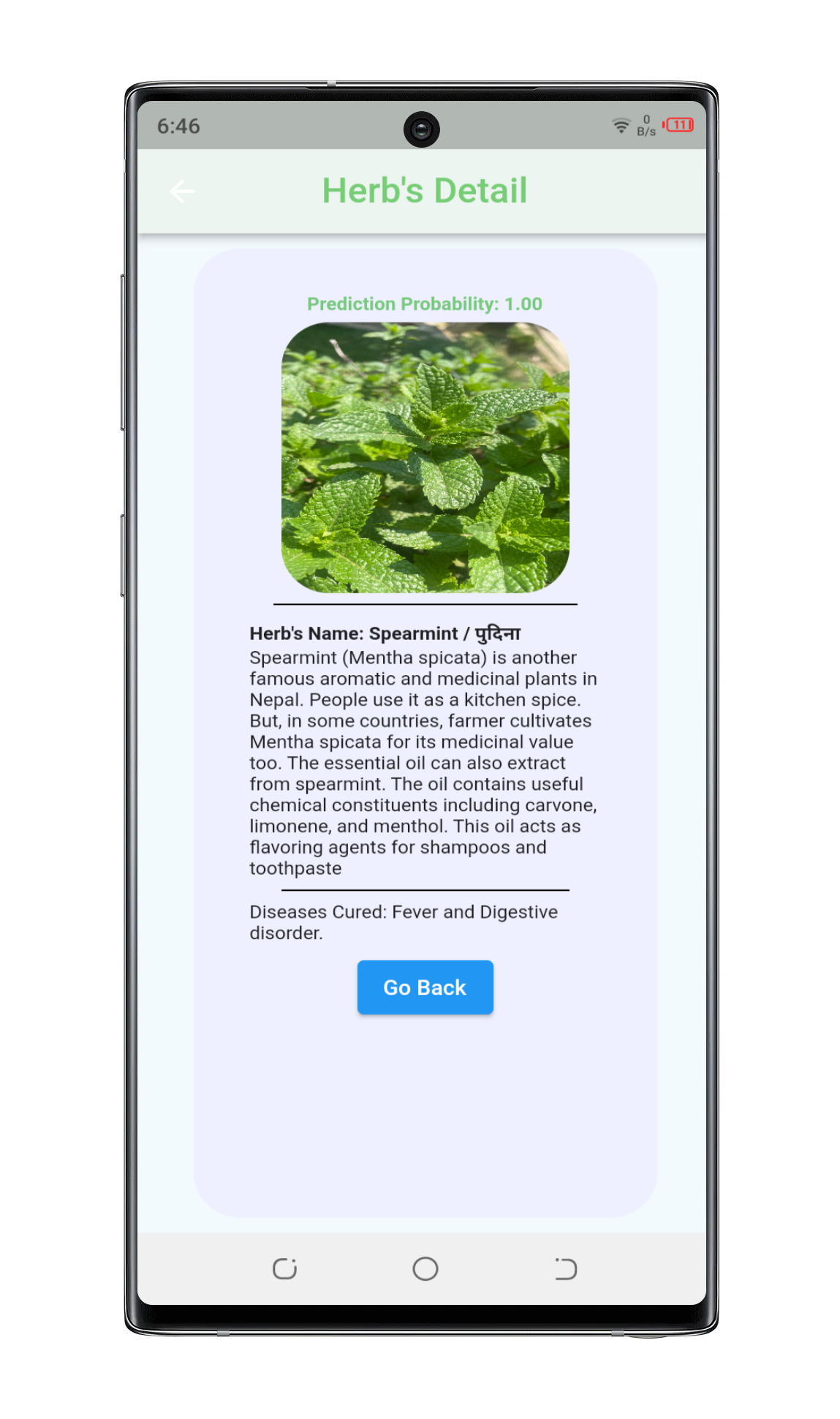}
    \end{minipage}
    \caption{Screenshots of the mobile application for herb classification.}
\end{figure}

\section{Conclusion}

The classification of 60 herbs opens up a wide range of possibilities for future use. One exciting application is its integration into mobile apps, allowing users to easily identify herbs and learn about their medicinal and culinary benefits. Developers can also build APIs around the system, making it easier to include herb-related features in other apps and platforms. With machine learning, this classification can support transfer learning to develop models for herb recognition and predicting potential uses. It also encourages collaboration between researchers, botanists, and herbalists to expand knowledge and validate traditional practices. E-commerce sites can benefit too, using the system for better product information and quality checks. In education, the classification can be used to teach people about herbs, while in research, it can help identify plants with unique properties. Finally, it can support conservation by guiding sustainable harvesting and protection efforts. In short, this system has the potential to make a meaningful impact across technology, science, commerce, and the environment.
%%%%%%%%%%%%%%%%%%%%%%%%%%%%%%%%%%%%%%%%%%%%%%%%%
\bibliographystyle{unsrt}
\bibliography{herbs}
%\nocite{*}

\end{document}